\documentclass[a4paper]{article}
\usepackage{ISCSLP2024}
\usepackage{CJK}

\usepackage{color,cite}

\usepackage{multirow}

\usepackage{ifthen}
\newboolean{blind}
\setboolean{blind}{false} 
\title{Bridging Speech and Text: Enhancing ASR with Pinyin-to-Character Pre-training in LLMs
\thanks{$^*$Equal Contribution, $^\dagger$Corresponding Author}
}
\name{
	\ifthenelse{\boolean{blind}}{Anonymous to ISCSLP}
	{Yang Yuhang$^{1*}$, Peng Yizhou$^{2*}$, Eng Siong Chng$^2$ and Xionghu Zhong$^1$$^\dagger$}
}

\address{
  \ifthenelse{\boolean{blind}}{Anonymous to ISCSLP}
  {
  	$^1$Hunan University, China\\
        $^2$Nanyang Technological University, Singapore
  }
}

\email{
	\ifthenelse{\boolean{blind}}{Anonymous to ISCSLP}
	{\{xzhong,yuhangear\}@hnu.edu.cn}
}

\begin{document}

\maketitle
\begin{abstract}
The integration of large language models (LLMs) with pre-trained speech models has opened up new avenues in automatic speech recognition (ASR). While LLMs excel in multimodal understanding tasks, effectively leveraging their capabilities for ASR remains a significant challenge. This paper presents a novel training approach to enhance LLM performance in ASR tasks. We propose pre-training LLMs on Pinyin embedding sequences, which represent pronunciation features, to generate corresponding Chinese characters. This step enables the LLM to adapt to generating text from pronunciation features before encountering real speech data. Furthermore, we fine-tune the LoRA parameters to enhance the LLM's understanding of speech modality information. In AISHELL-1 corpus, our approach yields a 9.5\% relative improvement in ASR tasks compared to the baseline without Pinyi-to-Character pre-training. Additionally, incorporating auxiliary text data for Pinyi-to-Character pre-training further boosts performance, achieving a 19.0\% relative improvement.

\end{abstract}

\section{Introduction}

LLMs trained on vast amounts of data exhibit remarkable semantic understanding apabilities~\cite{mann2020language,chowdhery2023palm,le2023bloom,qwen,llama2023,Gpt_4}. Through fine-tuning on downstream tasks, they excel in various domains such as abstractive summarization, question answering, knowledge retrieval, text generation, and machine translation~\cite{llm_survey}. In the field of ASR, effective models require two key components: robust audio feature extraction and semantic comprehension. The latter is crucial for addressing challenges such as homophones and poor audio quality. Integrating the powerful semantic understanding of LLMs into ASR systems has thus become a focal point of current research.

Speech pre-training models have made significant progress, leveraging large audio datasets to extract high-quality features and achieve impressive performance in downstream tasks. Remarkable capabilities in extracting speech features have been demonstrated by several models (e.g.,\cite{wav2vec,chen2022wavlm,hsu2021hubert,whisper,chen2024eat,shi2023multi}). The current stage is ripe for combining the powerful semantic understanding of LLMs with pre-trained speech models, which is expected to boost ASR recognition performance to a new level.

Recent advancements have demonstrated the potential of LLMs in multimodal applications \cite{Blip_2,liu2024visual,gao2023llama,su2023pandagpt,lyu2023macaw,llama_whispering}. Notably, a lightweight Querying Transformer was introduced to address the misalignment between visual and textual feature spaces, thereby enhancing the multimodal capabilities of LLMs \cite{Blip_2}. Furthermore, a frozen text-to-image diffusion model\cite{image_llm_1} is proposed, which learns the mapping between the LLM's embeddings and the diffusion model to enable text-to-image generation. Additionally, the ability to perceive and generate outputs from arbitrary combinations of text, images, videos, and audio inputs has been demonstrated, showcasing the versatility of LLMs in handling diverse multimodal inputs \cite{next_gpt}. 

In the realm of audio applications, LLMs have been increasingly integrated into ASR systems, showcasing innovative approaches and methodologies. When incorporating speech features into LLMs, two crucial issues must be considered: the mismatch between speech and text lengths and the inconsistency between feature modalities. Speech features with excessive lengths can hinder the alignment between text and audio in attention mechanisms, leading to out-of-vocabulary (OOV) problems. Various approaches have been proposed to address these issues. For example, convolutional networks are used to reduce the original Conformer model output length to one-third, feeding the speech features as prompts into the LLM for ASR training \cite{prompt_asr}. The CTC Compressor directly compresses audio feature length to text feature length \cite{one_decoder}. A Q-Former based approach with a fixed number of output frames is adopted \cite{Salmonn,connect_asr_encoder}, while a simple average pooling method is applied to reduce feature length \cite{chu2023qwen}. Additionally, concatenating adjacent features and passing them through a linear layer is another method used to reduce feature length \cite{embarrassingly}.

Despite the significant differences between speech and text modalities, many experiments have achieved promising results using various methods to address these differences. For instance, discrete representations are directly extracted from pre-trained audio encoders to enhance LLMs with cross-modal conversational abilities through a three-stage training process \cite{speechgpt}. In another approach, the LLM model is frozen, and gradients are passed to the audio encoder, transforming it into a semantic-rich, LLM-required audio encoder \cite{chu2023qwen}. Additionally, A simple linear projection achieves audio-to-text feature conversion when both the LLM and the audio encoder are frozen\cite{embarrassingly}. Maximizing the preservation of speech information extracted by the audio encoder and enabling LLMs to understand semantics is a valuable research direction.

Previous research has mainly concentrated on model architecture design and the incorporation of speech features into LLMs. However, the crucial aspect of how LLMs transcribe these features has been underexplored. We are motivated by the need for LLMs to understand speech modality information and convert pronunciation sequences into transcription text. To achieve this, we propose a novel two-stage training method. First, we pre-train LLMs using a Pinyin-to-character task, which enables them to convert Pinyin embedding sequences into corresponding Chinese characters. Here, Pinyin embeddings can represent pronunciation features. After LLMs learn to convert Pinyin features into corresponding text, we fine-tune the model parameters to make them understand genuine speech features and output the corresponding text sequences.

Our contributions in this paper are twofold: 1) We propose a novel pre-training method for LLMs, where we utilize Pinyin embedding sequences as a bridge between speech and text modalities. By converting Pinyin sequences into corresponding Chinese characters, we enable LLMs to adapt to generating text based on pronunciation before encountering real speech features, thereby enhancing their ability to comprehend speech modality information. 2) We show that incorporating additional pure text data during training can further improve the performance of LLM-based ASR models, which is particularly beneficial for low-resource ASR tasks.

\section{Method}

To enhance the performance of LLMs in ASR, we propose a two-stage training approach. 
In the first stage, we fine-tune the LLM on a task that predicts Chinese characters from Pinyin inputs. This task enables the LLM to adapt to converting phonetic units to text, as Pinyin closely represents pronunciation. This stage not only transforms the LLM from a next-token prediction model to one that can interpret other feature inputs but also leverages abundant, cost-effective text-only data, increasing the model's adaptability in the target domain without requiring paired speech data.

In the second stage, we utilize a pre-trained audio model to extract audio features from audio data. These features are then downsampled in order to reduce their length and then fed into the LLM model. By training the model using the LoRA~\cite{lora} method, which updates a low-rank subset of parameters, we enable the model to adapt to accept speech features and predict the corresponding transcriptions.

\subsection{Fine-tuning on Pinyin-to-Character Prediction}
Although LLMs have been trained on vast amounts of text data and possess strong semantic understanding capabilities, they still lack the ability to predict corresponding text based on pronunciation. This limitation can be addressed by fine-tuning the model on Pinyin-to-Character data, which is particularly important for low-resource or special-domain data that may not have been adequately represented in the LLM's initial training.

Pinyin serves as a bridge between spoken Chinese and its written form. By training the model to predict characters from Pinyin inputs, we can leverage a large amount of available textual data to improve the model's performance in ASR tasks without requiring audio data initially.

\begin{figure}[t]
  \centering
  \includegraphics[width=\linewidth]{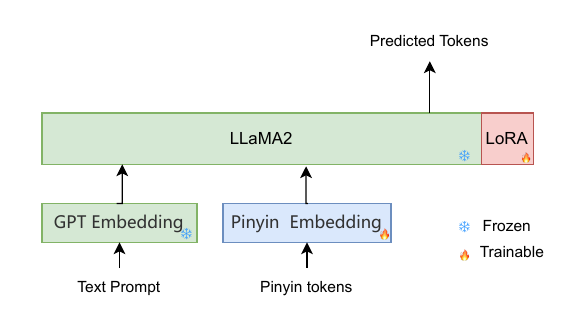}
  \caption{Fine-tuning on Pinyin-to-Character Prediction.}
  \label{fig:speech_production}
\end{figure}

As illustrated in Figure~\ref{fig:speech_production}, we augment the LLaMA2 model with additional Pinyin embeddings as inputs. The model employs LoRA for fine-tuning, which enhances the model's adaptability by introducing low-rank modifications to its weights, allowing for efficient fine-tuning with fewer parameters. The inputs to the model consist of a text prompt embedding $\mathbf{E}_t$ and a Pinyin embedding $\mathbf{E}_p$, and the output is the corresponding Chinese text.

Let $\mathbf{T} = (t_1, t_2, \ldots, t_K)$ be the text prompt and $\mathbf{P} = (p_1, p_2, \ldots, p_T)$ be the Pinyin sequence. The generation of the corresponding Chinese text sequence $\mathbf{Y} = (y_0, y_1, \ldots, y_{M-1})$ with the LLaMA2 model can be formulated as:

\begin{equation}
p\left(\mathbf{Y} \mid \mathbf{T}, \mathbf{P} ; \Theta_{\mathrm{LLM}}\right)=\prod_{m=0}^{M-1} p\left(y_{m} \mid \mathbf{Y}_{<m}, \mathbf{E}_{t}, \mathbf{E}_{p} ; \Theta_{\mathrm{LLM}}\right)
\end{equation}

where $\mathbf{Y}_{<m} = (y_0, y_1, \ldots, y_{m-1})$ indicates the generated text sequence before $y_m$, and $\Theta_{\text{LLM}}$ represents the model parameters.

During the training process, we freeze the parameters of the LLM and the GPT embedding layer and only update the parameters of the LoRA adapter and the Pinyin embedding layer. This approach allows us to adapt the model to the Pinyin-to-character prediction task.

To enhance the Large Language Model's Pinyin-to-character conversion capability, we design three tasks that foster semantic understanding beyond mere phonetic mapping. As shown in Table~\ref{tab:multi_task}, the three tasks have different text prompts. In the table, the Pinyin Input includes a sequence with tones, while the sequence without tones, one of the situations shown in our experiments, is "hen duo ji qi ren dou neng xia qi".

These tasks are designed to challenge the LLM to develop a deeper understanding of the relationships between Pinyin inputs and their corresponding Chinese character outputs.
\begin{CJK}{UTF8}{gbsn}

\begin{table}[th]
  \caption{ Multi-Prompt Tasks}
  \label{tab:multi_task}
  \centering
  \begin{tabular}{ p{10pt} p{57pt} p{58pt}  p{60pt} }
    \toprule
    
     \centering \textbf{Task} & \centering \textbf{Text Prompt} & \textbf{Pinyin Input}  & \textbf{Output Tokens} \\
    \midrule
    
    \centering \textbf{1} & 将语音特征转换成中文序列 & hen3 duo1 ji1 qi4 ren2 dou1 neng2 xia4 qi2 & 很多机器人都能下棋 \\
    \midrule
    \centering \textbf{2} & 将缺失的语音特征转换成中文序列 & hen3 duo1 ji1 [mask] ren2 dou1 neng2 xia4 qi2 & 很多机器人都能下棋 \\
    \midrule
    \centering \textbf{3} & 标记出同音字错误的地方 & 鸡，棋 ,hen3 duo1 ji1 qi4 ren2 dou1 neng2 xia4 qi2 & 很多[机]器人都能下棋 \\
    
    \bottomrule
  \end{tabular}
\end{table}

By tackling these tasks, the LLM develops a more profound understanding of contextual relationships and sequence structures, ultimately enabling it to accurately transcribe spoken language and capture its subtleties. 

The first task, "Converting Speech Features to Chinese Characters" uses the prompt "将语音特征转换成中文序列" and takes Pinyin embeddings as input. The LLM is tasked with predicting the corresponding Chinese characters.
The second task, "Converting Incomplete Speech Features to Chinese Characters" uses the prompt "将缺失的语音特征转换成中文序列" and involves Pinyin embeddings that are masked with a probability of 0.2. Given the randomly masked input embeddings, the LLM is expected to predict the corresponding Chinese characters correctly. The third task, "Identifying Homophone Errors" uses the prompt "标记出同音字错误的地方" and concatenates embeddings for homophone characters extracted from the GPT embedding layer before the Pinyin embeddings in the input sequence, with the goal of identifying and marking the positions of homophone errors.

\end{CJK}

\subsection{Fine-tuning on Speech-to-Character Prediction.}

\begin{figure}[t]
  \centering
  \includegraphics[width=\linewidth]{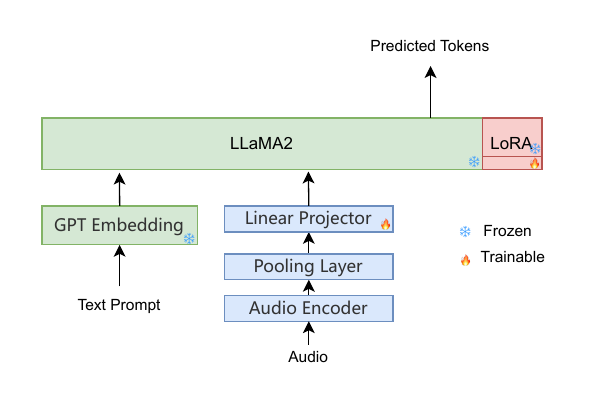}
  \caption{Schematic diagram of speech production.}
  \label{fig:speech_production2}
\end{figure}

After pre-training on the Pinyin-to-character task, the LLM has acquired the ability to convert Pinyin embeddings into Chinese characters. However, the model has not been exposed to real acoustic features. In the second stage of training, we aim to integrate acoustic features into the LLM while minimizing the increase in model parameters and reducing information loss.

As illustrated in Figure~\ref{fig:speech_production2}, regarding the audio encoder, we adopt a hybrid feature reduction approach, combining pooling method from Qwen-audio~\cite{chu2023qwen}  
and the linear projector from~\cite{embarrassingly} 
to reduce the dimensionality of acoustic features. This hybrid approach aims to avoid the drawbacks of using either method alone. Specifically, the pooling method would average out distinctive semantic features, contradicting the short-time stationarity of audio signals, while the linear projector would exponentially increase the input dimension of the linear layer. Specifically, we first apply a pooling layer to average adjacent acoustic features, followed by concatenating the averaged features. Then, a single linear layer is used to transform the acoustic features into the same dimensionality as the GPT embeddings, which is 4096. This process can be denoted as follows:

\[
\mathbf{H}_S = \text{Encoder}(\mathbf{X}_S)
\]
\[
\mathbf{E}_S = \text{Linear}(\text{pooling}(\mathbf{H}_S))
\]

where $\mathbf{X}_S = (x_1, x_2, \ldots, x_N)$ denotes the speech input, $\mathbf{H}_S = (h_1, h_2, \ldots, h_M)$ represents the hidden embeddings, and $\mathbf{E}_S = (e_1, e_2, \ldots, e_T)$ represents the transformed acoustic features.

Given the text prompt $\mathbf{T} = (t_1, t_2, \ldots, t_K)$ and speech input $\mathbf{E_S}$, the generation of the corresponding Chinese text sequence $\mathbf{Y} = (y_0, y_1, \ldots, y_{M-1})$ with the LLaMA2 model can be formulated as:
\begin{equation}
    p\left(\mathbf{Y} \mid \mathbf{T}, \mathbf{E}_{S} ; \Theta_{\mathrm{LLM}}\right)=\prod_{m=0}^{M-1} p\left(y_{m} \mid \mathbf{Y}_{<m}, \mathbf{E}_{t}, \mathbf{E}_{S} ; \Theta_{\mathrm{LLM}}\right)
\end{equation}

During multimodal fusion training, we update only the linear layer and the lower layers of the LLM while freezing the higher layers to prevent the loss of previously learned Pinyin-to-character abilities.

\section{Experimental setup}

\subsection{Data}
In our experiments, we leverage the AISHELL-1 dataset~\cite{AISHELL-1}, a widely acknowledged corpus in the ASR research community. The training set of the AISHELL-1 dataset comprises approximately 120K utterances, totaling around 150 hours of speech data. 
The development set and test set consist of around 10 and 5 hours of speech data, respectively.
For the Pinyin-to-character tasks, we extracted 100 times the amount of pure text from the Wenetspeech~\cite{zhang2022wenetspeech} dataset for text pre-training.

\subsection{Modeling}

 The architecture consists of three main components: the Large Language Model, the pre-trained speech model, and the feature transformation module. We employ LLaMA2 as our LLM and Whisper-large-v3 as the speech encoder.

The Whisper-large-v3 model, trained for speech recognition and translation on a large amount of weakly supervised data, generates encoder output features that are well-suited for modeling speech and capturing information about background noises. The LLaMA2-7b model~\cite{gao2023llama} we utilize in our implementation offers advanced language understanding capabilities and can handle complex language tasks effectively.

For feature transformation, we extract features from Whisper-large-v3 and apply an average pooling layer with a kernel size of 3 and a stride of 3 to reduce the dimension of the speech features by a factor of 3. Subsequently, we concatenate adjacent speech features and pass them through a linear layer with an output dimension of $(1280 \times 3, 4096)$, which transforms the speech features to a dimension that matches the Pinyin embeddings.
Our experiments make use of the LLaMA2-Accessory toolkit\footnote{LLaMA2-Accessory: https://github.com/Alpha-VLLM/LLaMA2-Accessory}, an open-source framework for developing large language models.

For the Pinyin-to-character task, we generate Pinyins using the python-pinyin toolkit\footnote{python-pinyin: https://github.com/mozillazg/python-pinyin}. The Pinyin embeddings, having the same dimension as the LLM embeddings, that is, 4096, are derived from 1417 Pinyin units in all characters shown in the Aishell-1 dataset. We tested various configurations, including the multi-task setup described in Table \ref{tab:multi_task}. Additionally, we explored the impact of tone on performance by using Pinyins without tones and directly embedding Pinyins using GPT embeddings. We also investigated the effect of text data quantity on performance, finding that increasing the amount of text data improved the results. Specifically, we used 120,000 text data samples.

For the speech-to-character task, we downsample the audio data to one-ninth of the original size and feed it into the LLM. We trained the lowest 5 layers of LoRA parameters in the LLM for multimodal fusion. 

\subsection{Training Details}

In the first stage, we fine-tune the LLaMA2 model using LoRA adapters to the Pinyin-to-character task. We set the LoRA rank to 16 and enable bias and norm tuning. The learning rate is initialized at $1 \times 10^{-4}$, with a linear warmup period of one epoch, followed by a decay to a minimum learning rate of $5 \times 10^{-6}$. The AdamW optimizer is used with a batch size of 60. The model is initially pre-trained on specific multi-prompt tasks for 10 epochs, after which it is fine-tuned on task 1 specified in Table~\ref{tab:multi_task} for an additional 2 epochs.

In the second stage, we address the speech-to-character task by training on speech data. To reduce computational costs, we down-sample the original speech features to one-ninth of their original size before inputting them into the LLM. During this stage, the focus is on training the linear layer in the down-sampling component and the LoRA parameters of the first 5 layers of the LLM. We maintain the same hyperparameters as in the first stage, including a learning rate, warmup steps, and batch size.

\section{Results}

\subsection{Pinyin-to-Character Tasks}

    
    
    

\begin{table}[th]
   \caption{CER for Various Pinyin-to-Character Tasks. \textbf{SysID} means System ID. \textbf{Tone} indicates whether tone is included. \textbf{Emb} refers to the embedding type. \textbf{PY} stands for pinyin embeddings, and \textbf{GPT} stands for GPT embeddings. System 7 uses 100 times more text data based on the same configuration as System 6.}
  \label{tab:wer_syllable_to_char}
  \centering
  \begin{tabular}{p{20pt} p{20pt} p{50pt} p{20pt} | cc  }
    \toprule
    \multicolumn{4}{c}{\textbf{Configs}} & \multicolumn{2}{|c}{\textbf{TestSets}} \\ \midrule
    \textbf{SysID} & \textbf{Tone} & \textbf{Prompt Task} & \textbf{Emb} & \textbf{Dev} & \textbf{Test} \\ \hline

  1 & $\times$   & Task1 & PY & 7.3 & 7.2 \\ 
  2 & \checkmark & Task1 & GPT & 3.6 & 3.6 \\ 
  3 & \checkmark & Task1 & PY &  3.0 & 2.9 \\ 
  4 & \checkmark & Task2 & PY & 2.7 & 2.6 \\ 
  5 & \checkmark & Task3 & PY & 2.5 & 2.6 \\ \hline
  6 & \checkmark & Task2\&Task3  & PY & 2.3  & 2.4  \\  
  7 & \multicolumn{3}{c|}{ ~~~~ + 100x text data}  & 2.0 & 1.9 \\
  \bottomrule

  \end{tabular}
\end{table}

Table \ref{tab:wer_syllable_to_char} presents the Character Error Rates (CERs) for the Pinyin-to-character task across various experimental setups. We first examined the Pinyin-to-character task without tonal information, shown as System 1, achieving CERs of 7.3\% on the dev set and 7.2\% on the test set. This task posed a significant challenge due to the inherent ambiguity in tonal languages when tones are omitted. Subsequently, we incorporated tonal information into the Pinyin sequences during training. We utilize the built-in GPT embedding layer of LLaMA for Pinyin encoding, comparing it against a newly designed Pinyin embedding, as shown in Systems 2 and 3. The results demonstrated that our Pinyin embedding significantly outperformed the generic GPT tokenizer, which does not intuitively capture the nuances of Pinyin sequences. This enhancement resulted in a reduction of CER to 3.0\% on the dev set and 2.9\% on the test set. 

We further explored the impact of Task2 and Task3 mentioned in Table~\ref{tab:multi_task}, which resulted in System 4 and System 5. These approaches further reduced CER to 2.7\% and 2.5\% on the dev set and 2.6\% on the test set, respectively, resulting in a relative performance improvement of over 10\%. System 6 combines Task2 and Task3, resulting in a CER of 2.3\% on the dev set and 2.4\% on the test set, representing a relative improvement of 23\% and 17\%, respectively, compared to System 3.

Finally, we use 100 times the amount of text following the same configuration as System 6, resulting in System 7. This strategy further reduced the CER to 2.0\% on the development set and 1.9\% on the test set, demonstrating the benefits of large-scale training data in enhancing model performance.

\subsection{Speech-to-Character Tasks}

\begin{table}[th]
  \caption{CER for Various Speech-to-Character Tasks with Text Pretraining. System 6 and System 7 from Table~\ref{tab:wer_syllable_to_char} are used as initial LLMs here}
  \label{tab:wer_speech_to_char}
  \centering
  \begin{tabular}{ p{20pt} p{60pt} | cc }
    \toprule
    \multicolumn{2}{c}{\textbf{Configs}} & \multicolumn{2}{|c}{\textbf{TestSets}} \\ 
    \midrule
    \textbf{SysID} & \textbf{LLMs source}  & \textbf{Dev}  & \textbf{Test} \\
    \midrule
    \hspace{3pt}8 & LLaMA2 & 6.6 & 6.3 \\
    \midrule
    \hspace{3pt}9 & System 6  & 5.9 & 5.7 \\
    10 & System 7 & 5.3  & 5.1 \\
    \bottomrule
  \end{tabular}
\end{table}

The results presented in Table~\ref{tab:wer_speech_to_char} highlight the significant impact of text pretraining on the performance of the speech-to-character recognition task. In System 8, the LLM is directly initialed from the LLaMA2-7b model without text pertaining to the Pinyin-to-character task. The CER result is 6.6\% on the dev set and 6.3\% on the test set. 

When introducing text pertaining, performing System 9, which uses System 6 as pretrained LLM, remarkably enhances performance, reducing the CER by 10.6\% on the dev set and 9.5\% on the test set relatively. Furthermore, System 10 scales up the dataset for text pretraining to 100 times, using the LLM from System 7, leading to an additional improvement, lowering the CER to 5.3\% on the dev set and 5.1\% on the test set. The resulting final relative improvements achieved are 19.6\% on the dev set and 19.0\% on the test set. These results demonstrate that extensive text pretraining significantly enhances ASR system performance.

\section{Discussion \& Conclusion}\label{sec:discussion}


In this study, we introduced a novel two-stage training approach to enhance automatic speech recognition (ASR) systems by integrating large language models (LLMs) with pre-trained speech models. Pre-training the LLMs on Pinyin-to-character tasks enabled the model to generate text from pronunciation features before encountering speech data, significantly improving its semantic comprehension. This approach yielded a 9.5\% relative improvement in ASR tasks for AISHELL-1 corpus, with further gains up to 19.0\% when auxiliary text data was incorporated. Our method effectively bridges the gap between speech and text modalities, making it particularly beneficial for low-resource ASR tasks.


Despite the progress made, the current experimental results are not ideal. One significant limitation is that the powerful text generation capabilities of LLMs can lead to overfitting, especially when the training data is limited. Furthermore, the current approach to integrating speech models with LLMs may not be optimal for smaller datasets. In future work, we plan to expand the training dataset and explore models that perform better on Chinese tasks, such as the Qwen model~\cite{qwen}. Additionally, we aim to further refine the integration of pre-trained audio encoders with LLMs to optimize performance.

\bibliographystyle{IEEEtran}

\bibliography{mybib}

\end{document}